\documentclass[runningheads]{llncs}

\usepackage[mobile]{eccv}

\usepackage{xcolor}         

\usepackage[dvipsnames,svgnames,x11names]{xcolor}
\usepackage{tikz}
\usetikzlibrary{arrows.meta,shapes,calc,matrix,fit,positioning,backgrounds,decorations.markings}
\usepackage{pgfplots}
\usepackage{pgfplotstable}
\pgfplotsset{compat=1.9}
\usepackage{xstring}

\usepgfplotslibrary{external}

\IfBeginWith*{\jobname}{fig/extern/}{\finalcopy}{}

\tikzstyle{every picture}+=[
	remember picture,
	every text node part/.style={align=center},
	every matrix/.append style={ampersand replacement=\&},
]
\tikzstyle{tight} = [inner sep=0pt,outer sep=0pt]
\tikzstyle{node}  = [draw,circle,tight,minimum size=12pt,anchor=center]
\tikzstyle{op}    = [draw,circle,tight]
\tikzstyle{dot}   = [fill,draw,circle,inner sep=1pt,outer sep=0]
\tikzstyle{pt}    = [fill,draw,circle,inner sep=1.5pt,outer sep=.2pt]
\tikzstyle{box}   = [draw,rectangle,inner sep=3pt]
\tikzstyle{high}  = [black!60]
\tikzstyle{group} = [high,box,opacity=.5]
\tikzstyle{dim1}  = [fill opacity=.3,text opacity=1]
\tikzstyle{dim2}  = [fill opacity=.5,text opacity=1]
\tikzstyle{dim3}  = [fill opacity=.7,text opacity=1]
\tikzstyle{rectc} = [tight,transform shape]
\tikzstyle{rect}  = [rectc,anchor=south west]

\tikzset{every mark/.append style={solid}}
\pgfplotsset{
	grid=both, width=\columnwidth, try min ticks=5,
	every axis/.append style={font=\small},
	every axis plot/.append style={thick,mark=none,mark size=1.8,tension=0.18},
	legend cell align=left, legend style={fill opacity=0.8},
	xticklabel={\pgfmathprintnumber[assume math mode=true]{\tick}},
	yticklabel={\pgfmathprintnumber[assume math mode=true]{\tick}},
	nodes near coords math/.style={
		nodes near coords={\pgfmathprintnumber[assume math mode=true]{\pgfplotspointmeta}},
	},
}

\pgfplotsset{
	dash/.style={mark=o,dashed,opacity=0.6},
	dott/.style={mark=o,dotted,opacity=0.6},
	nolim/.style={enlargelimits=false},
	plain/.style={every axis plot/.append style={},nolim,grid=none},
}

\pgfdeclarelayer{bg4}
\pgfdeclarelayer{bg3}
\pgfdeclarelayer{bg2}
\pgfdeclarelayer{bg1}
\pgfdeclarelayer{fg1}
\pgfdeclarelayer{fg2}
\pgfdeclarelayer{fg3}
\pgfdeclarelayer{fg4}
\pgfsetlayers{bg4,bg3,bg2,bg1,main,fg1,fg2,fg3,fg4}

\pgfkeys{/tikz/.cd, aspect/.store in=\aspect, aspect=1}
\pgfkeys{/tikz/.cd, depth/.store in=\depth, depth=.5}
\pgfkeys{/tikz/.cd, stepx/.store in=\step, stepx=1}

\tikzstyle{geom} = [line join=bevel,aspect=1,depth=.5,z={(\depth*\aspect,\depth)}]
\tikzstyle{wire} = [geom,draw,thick]

\def\cx[#1,#2,#3]{#1}
\def\cy[#1,#2,#3]{#2}
\def\cz[#1,#2,#3]{#3}
\def\ex[#1,#2,#3]{#1,0,0}
\def\ey[#1,#2,#3]{0,#2,0}
\def\ez[#1,#2,#3]{0,0,#3}

\usepackage[export]{adjustbox}

\newcommand{\Th}[1]{\textsc{#1}}
\newcommand{\mr}[2]{\multirow{#1}{*}{#2}}

\newcommand{\red}[1]{{\textcolor{red}{#1}}}

\newcommand{\citeme}[1]{\red{[XX]}}
\newcommand{\refme}[1]{\red{(XX)}}

\makeatletter
\newcommand*\bdot{\mathpalette\bdot@{.7}}
\newcommand*\bdot@[2]{\mathbin{\vcenter{\hbox{\scalebox{#2}{$\m@th#1\bullet$}}}}}
\makeatother

\makeatletter
\DeclareRobustCommand\onedot{\futurelet\@let@token\@onedot}
\def\@onedot{\ifx\@let@token.\else.\null\fi\xspace}

\makeatother
\newcommand{\graycell}[1]{\textcolor{gray}{#1}}

\newcommand{\ours}{\texttt{Re2Pix}\xspace}

\definecolor{ForestGreen}{RGB}{34,139,34}
\definecolor{Orange}{RGB}{1.0, 0.55, 0.0}
\definecolor{Purple}{RGB}{0.58, 0.44, 0.86}
\definecolor{softblue}{rgb}{0.2, 0.4, 0.8} 
\definecolor{coral}{RGB}{255, 127, 80} 
\definecolor{softcoral}{rgb}{1.00, 0.60, 0.50}  

\definecolor{teal}{rgb}{0.1, 0.6, 0.6} 
\definecolor{forestgreen}{rgb}{0.1, 0.6, 0.2} 

\definecolor{vibrantorange}{rgb}{1.0, 0.5, 0.0}

\definecolor{deepblue}{rgb}{0.0, 0.0, 0.8}

\definecolor{TableColor}{rgb}{0.58, 0.55, 0.51}

\definecolor{ForestGreen}{RGB}{34,139,34}
\definecolor{Orange}{RGB}{1.0, 0.55, 0.0}
\definecolor{Purple}{RGB}{0.58, 0.44, 0.86}
\definecolor{VibrantMagenta}{RGB}{255,0,197}
\definecolor{DarkenedMagenta}{RGB}{225,0,167}

\definecolor{TableColor}{rgb}{0.882, 0.945, 0.882}

\newcolumntype{C}[1]{>{\centering\arraybackslash}p{#1}}

\def\addlegendimage{\csname pgfplots@addlegendimage\endcsname}

\newcommand{\secref}[1]{\hyperref[#1]{Sec.~\ref{#1}}}
\newcommand{\sectionref}[1]{\hyperref[#1]{Section~\ref{#1}}}

\newcommand{\supplautoref}[1]{\hyperref[#1]{Suppl.~\autoref{#1}}}
\newcommand{\supplsec}[1]{\hyperref[#1]{Suppl. Sec.~\ref{#1}}}

\usepackage{multirow}
\usepackage{enumitem}
\usepackage[normalem]{ulem}
\usepackage{xspace}
\usepackage{amsmath}

\usepackage{tikz}
\usepackage{pgfplots}
\usepackage{xspace}
\usepackage{amsmath}
\usepackage{colortbl}
\usepackage{enumitem}
\usepackage[normalem]{ulem}
\usepackage{wrapfig}
\usepackage{caption} 
\usepackage{graphicx}
\usepackage{multirow}
\usepackage{etoc}

\usepackage{eccvabbrv}

\usepackage{graphicx}
\usepackage{booktabs}

\usepackage[accsupp]{axessibility}  

\usepackage[pagebackref,breaklinks,colorlinks,citecolor=eccvblue]{hyperref}

\usepackage{orcidlink}

\begin{document}

\title{Representations Before Pixels: Semantics-Guided Hierarchical Video Prediction} 

\titlerunning{Re2Pix}


\authorrunning{E.~Karypidis et al.}


\author{Efstathios Karypidis$^{1,3}$ \hspace{0.5em}Spyros~Gidaris$^{2}$ \hspace{0.5em}  Nikos~Komodakis$^{1,4,5}$ \vspace{0.5em} 
\\
$^1$Archimedes, Athena Research Center, Greece \hspace{1.0em} $^2$valeo.ai \\
$^3$National Technical University of Athens \hspace{1.0em} $^4$University of Crete \hspace{1.0em} $^5$IACM-Forth
}

\maketitle

\begin{abstract}
Accurate future video prediction requires both high visual fidelity and consistent scene semantics, particularly in complex dynamic environments such as autonomous driving. We present \ours, a hierarchical video prediction framework that decomposes forecasting into two stages: semantic representation prediction and representation-guided visual synthesis. Instead of directly predicting future RGB frames, our approach first forecasts future scene structure in the feature space of a frozen vision foundation model, and then conditions a latent diffusion model on these predicted representations to render photorealistic frames. This decomposition enables the model to focus first on scene dynamics and then on appearance generation. A key challenge arises from the train–test mismatch between ground-truth representations available during training and predicted ones used at inference. To address this, we introduce two conditioning strategies, nested dropout and mixed supervision, that improve robustness to imperfect autoregressive predictions. Experiments on challenging driving benchmarks demonstrate that the proposed semantics-first design significantly improves temporal semantic consistency, perceptual quality, and training efficiency compared to strong diffusion baselines. We provide the implementation code at \href{https://github.com/Sta8is/Re2Pix}{https://github.com/Sta8is/Re2Pix}.

\end{abstract}    
\section{Introduction}
\label{sec:intro}

Video prediction plays a central role in autonomous systems, where anticipating how a scene will evolve is essential for long-horizon reasoning and decision making~\cite{dosovitskiy2022learning,guan2024world,feng2025survey}. 
In driving scenarios, the ability to accurately forecast how visual scenes evolve, from the motion of vehicles and pedestrians to the subtleties of lighting and occlusions, is not merely a perceptual convenience but a prerequisite for safe planning~\cite{drive_wm,gao2024vista,tu2025role}. 
Yet learning such predictive models from raw video requires simultaneous mastery of high-level semantics (what objects are present and how they interact) and photorealistic details (how the scene appears) across temporal scales, a challenge that remains largely unsolved.

\begin{figure*}[t]
\centering
\includegraphics[trim={0cm 0cm 0cm 0cm},width=0.95\linewidth]{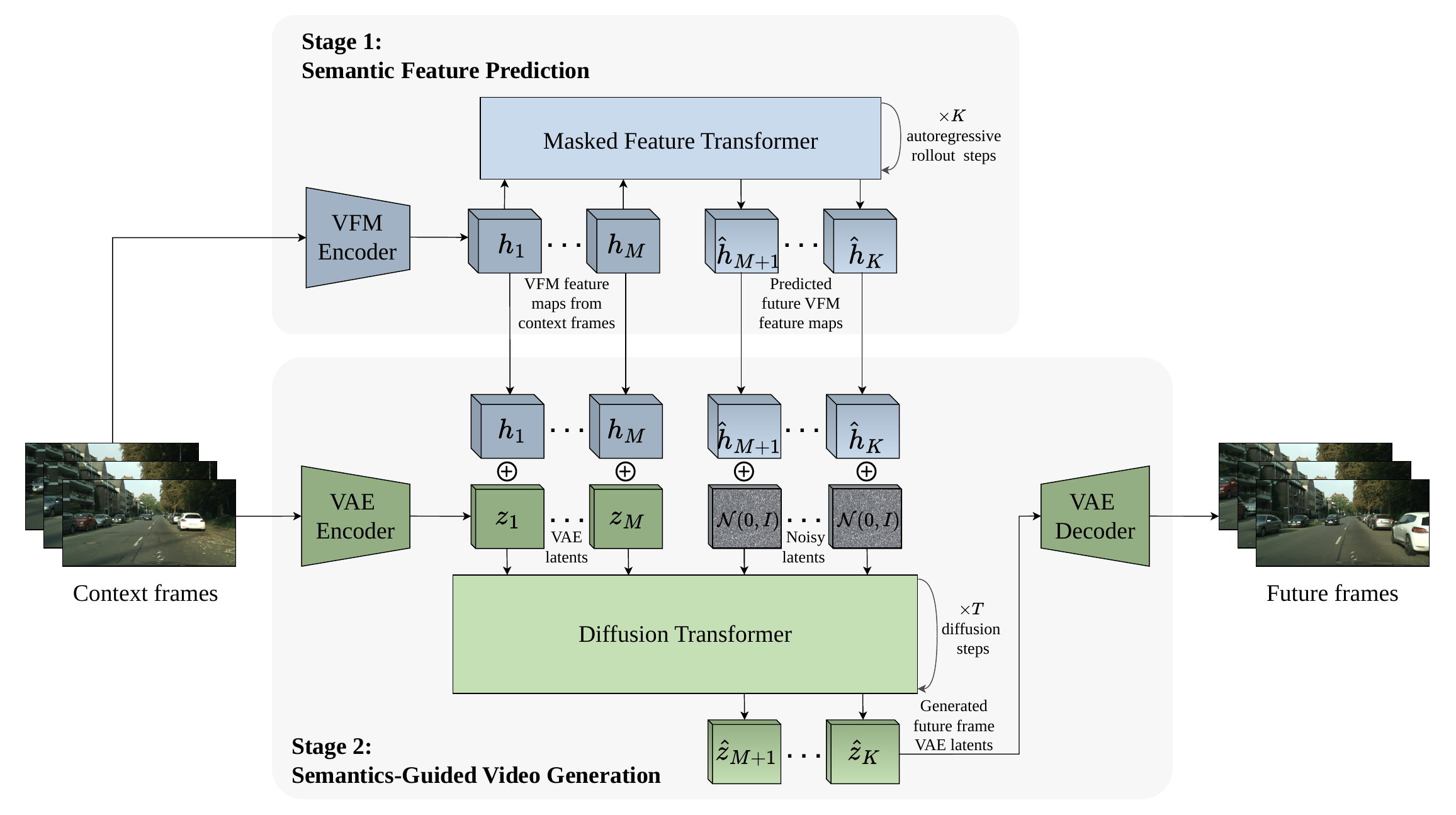}
\caption{\textbf{Overview of the proposed \ours hierarchical framework during inference.} 
In Stage 1, semantic features \( h_{1:M} \) of the context frames are extracted from a vision foundation model (VFM) encoder \(E_h\) and fed into a masked feature transformer to autoregressively predict (in frame-wise way) future semantic VFM features \( \hat{h}_{M+1:K} \). 
In Stage 2, both the past \( \hat{h}_{1:M} \) and predicted features \( \hat{h}_{M+1:K} \) condition the diffusion transformer \( G_z \) to generate future VAE latents \( \hat{z}_{M+1:K} \), 
which are then decoded into RGB future frames.}
\label{fig:overview}
\end{figure*}

Most modern approaches adopt an end-to-end paradigm: future frames are predicted directly in the latent space of a VAE, typically using diffusion models~\cite{ho2022imagen, peebles2023scalable}. 
While effective, this paradigm suffers from a fundamental limitation: semantic structure and fine-grained visual details are deeply entangled within the same latent representation. 
Consequently, the model must simultaneously infer scene dynamics and render photorealistic appearance, often leading to temporal semantic inconsistencies such as object identity drift, structural degradation, or flickering artifacts. 
More fundamentally, this entanglement slows convergence, increases data requirements, and makes it difficult to reason about or control each component independently.
Recent work has attempted to inject semantic structure into diffusion models by aligning intermediate features with pretrained representations via auxiliary distillation objectives~\cite{yu2024representation,zhang2025videorepa}. While such alignment can guide representation learning, the diffusion model still performs forecasting and rendering within a single latent space, leaving their roles implicitly coupled.

We ask a different question: \emph{can semantic forecasting and visual synthesis be explicitly separated, while still enabling coherent generative prediction?}

Our primary contribution is to introduce a hierarchical framework that decomposes video prediction into two interacting but distinct stages.
First, we forecast future scene structure in the feature space of a frozen vision foundation model (VFM). These representations capture high-level semantics while abstracting away low-level appearance. Second, we condition a latent diffusion model on the predicted representations to render photorealistic frames. This separation yields a structured predictive pipeline: the first stage models temporal dynamics in a representation space, while the second stage specializes in appearance synthesis conditioned on forecasted structure.

This clean separation, however, introduces a critical challenge. During training, the diffusion model has access to clean, ground-truth semantic features from future frames. At inference, it must instead rely on autoregressively predicted features, which inevitably accumulate errors. Naively supervising the generator solely on clean semantics creates a severe train–test mismatch: the model overfits to near-perfect conditioning signals and degrades sharply—often producing blurry or incoherent outputs—when exposed to noisy, imperfect forecasts. We demonstrate that bridging this distribution shift is essential for effective hierarchical video prediction.

Our second, technical contribution lies in modeling this bridge between semantic forecasting and generative synthesis. We adopt an early fusion strategy that token-wise merges VFM features with VAE latents at the input level, providing stable conditioning without increasing token count. To mitigate the conditioning mismatch, we introduce two complementary strategies. First, \emph{nested dropout} stochastically truncates feature channels during training, encouraging robustness to errors in the forecasted representation. Second, \emph{mixed supervision} exposes the generator to both ground-truth and predicted features (90/10 mixture), directly regularizing against over-reliance on idealized semantics. Together, these mechanisms enable the diffusion model to operate reliably under imperfect, autoregressively predicted features.

Concretely, our framework, \ours{}, extracts semantic representations using a frozen VFM encoder such as DINOv2~\cite{oquab2024dinov2}, and trains a lightweight masked Transformer to autoregressively predict future features in this space. A latent video diffusion model, implemented with a DiT backbone~\cite{peebles2023scalable} in the VAE latent space~\cite{wan2025wan}, is then conditioned directly on the \emph{forecasted} semantic features to render future frames. In contrast to alignment-based approaches such as REPA~\cite{yu2024representation} and VideoREPA\cite{zhang2025videorepa}, which encourage feature similarity between diffusion and semantic spaces, we treat forecasted VFM representations as an explicit intermediate generative variable.

\noindent\textbf{Our main contributions are:}
\begin{itemize} 
\item We introduce \ours, a hierarchical semantic-to-pixel framework that separates video forecasting into semantic representation prediction and semantics-driven visual generation. To our knowledge, we are the first to demonstrate that VFM feature prediction can effectively guide hierarchical video diffusion. 
\item We propose two robust conditioning strategies—\emph{nested dropout} and \emph{mixed supervision}—to close the train-test gap and improve robustness to imperfect, autoregressively generated features. 
\item Extensive results demonstrate that \ours achieves substantial improvements over strong baselines in temporal semantic fidelity and perceptual quality, while significantly accelerating training convergence (up to $7\times$ for generation and $14\times$ for segmentation metrics). 
\end{itemize}
\section{Related Work}
\label{sec:related}

\paragraph{Video Generation and Prediction.}
Video prediction has evolved from autoregressive models operating directly in pixel space~\cite{mathieu2016deep, Villegas2017MCNet, denton2018stochastic, Lee2018SAVP, wang2018eidetic, Gao_2022_CVPR} to hierarchical and structured approaches that model temporal dynamics more effectively~\cite{wang2018video, mallya2020world, yan2021videogpt, Ho2022VideoDiffusion, Yang2023LGCVD}. Transformer-based architectures have further advanced the field by leveraging autoregressive or masked modeling objectives to capture long-range temporal dependencies~\cite{yu2024language, yu2023magvit, gupta2023maskvit, wang2024emu3}. State-of-the-art video prediction systems typically operate in the latent space of a learned variational autoencoder~\cite{rombach2022high, kouzelis2025eq, wan2025wan}, where generative models predict future latent codes rather than raw pixels~\cite{gupta2023maskvit, yu2023magvit, yu2024language, gao2024vista, bartoccioni2025vavim}. Recent models trained on large-scale video data produce temporally coherent, photorealistic sequences~\cite{blattmann2023svd, He2022LatentVideo, polyak2024movie, yang2024cogvideox}. In controllable or world-modeling settings, Vista~\cite{gao2024vista} offers a generalizable driving world model with precise spatiotemporal control, while Cosmos-Predict~\cite{agarwal2025cosmos, cosmos_predict2, ali2025world} and Cosmos-Transfer~\cite{alhaija2025cosmos} enable multi-modal conditional generation for simulation and robotics.

Unlike these approaches, which predict future frames directly in pixel or VAE-latent space, our method introduces a hierarchical formulation that first forecasts high-level VFM semantic features and then generates pixels conditioned on them. This design improves temporal semantic consistency and reduces the burden on the diffusion model by providing stronger structural guidance. As a result, our framework bridges semantic forecasting with generative video modeling while maintaining competitive training efficiency.

\paragraph{Semantic Future Prediction.}
A line of work in future frame prediction focuses on forecasting semantic information rather than raw RGB values~\cite{nabavi2018future,sun2019predicting,jin2017video, vondrick2016anticipating}, typically predicting representations extracted from pre-trained networks. Early approaches~\cite{saric2020warp, hu2021apanet, lin2021predictive, karypidis2025advancing} targeted intermediate features or outputs of task-specific scene understanding models, such as Mask-RCNN~\cite{he2017mask} and Segmenter~\cite{strudel2021segmenter}.

More recently, DINO-Foresight~\cite{karypidis2024dino} and DINO-WM~\cite{zhou2025dinowm} forecast dense, patch-wise semantic from vision foundation models like DINOv2~\cite{oquab2024dinov2}, which, thanks to large-scale pre-training, generalize effectively to diverse tasks and new scenes without retraining. DINO-WM applies this approach to world modeling and action-conditioned planning in simulated environments, whereas DINO-Foresight focuses on multi-task dense semantic forecasting in real-world driving scenarios. Subsequent works extended this paradigm with diffusion-based formulations~\cite{walker2025generalist} and by scaling to larger datasets and models~\cite{baldassarre2025back}.

Relatedly, V-JEPA methods~\cite{bardes2024revisiting,assran2025v} learn visual representations by predicting masked video regions, but their objective is representation quality rather than forecasting future frames. In contrast, our work leverages VFM feature prediction as a hierarchical intermediate for RGB generation, enabling future frame synthesis with strong temporal semantic consistency while maintaining training efficiency.

\paragraph{Leveraging VFM features for visual generation.}
A growing body of work explores using features from pre-trained vision foundation models (VFMs)~\cite{oquab2024dinov2, tschannen2025siglip2, venkataramanan2025franca, simeoni2025dinov3} as strong priors for generative modeling. One line of methods aligns VAE latent spaces with VFM features through distillation losses~\cite{yao2025reconstruction,li2024imagefolder,chen2025masked, hu2023gaia, russell2025gaia}. Another aligns intermediate diffusion features with VFM representations~\cite{yu2024representation, leng2025repa}, an approach pioneered by REPA~\cite{yu2024representation}, which substantially accelerates diffusion training and improves image generation quality. These ideas have been extended to video generation~\cite{zhang2025videorepa, hwang2025cross, wu2025geometry}, where VFM-guided objectives improve temporal semantic consistency  and 3D geometry when \emph{fine-tuning} pretrained video diffusion transformers. However, these works do not demonstrate improved training convergence for video diffusion models trained \emph{from scratch}, nor do they address video prediction settings.

Recent work further leverages VFMs by jointly modeling low-level VAE latents and high-level VFM features within diffusion~\cite{kouzelis2025boosting, wu2025representation}, or by generating only high-dimensional VFM representations that are subsequently decoded to RGB~\cite{zheng2025diffusion}, both of which yield faster convergence and improved fidelity. Semantic representations have also been explored in hierarchical image synthesis pipelines that first predict global semantic representations and then generate VAE latents conditioned on them~\cite{pernias2024wurstchen, li2024return}.

In contrast to these approaches—which treat VFM features as static conditioning signals or as alternative generative latents—our method evolves VFM features through time, using them as a dynamic hierarchical intermediate for video generation. This semantics-guided forecasting enables temporally coherent and content-aware future frame synthesis, and, to our knowledge, we are the first to leverage VFM feature prediction for hierarchical video prediction.
\section{Methodology}
\label{sec:Method}

We address the video prediction task, where the goal is to forecast future frames given a sequence of past observations. Let \( x = (x_1, \ldots, x_K) \) be a video sequence of \( K \) frames. Given the first \( M \) frames (\( M < K \)), the task is to predict the remaining \( K-M \) frames. To solve this, we propose a hierarchical framework that decouples the problem into two stages (see \autoref{fig:overview}):
\begin{description}  
\item[High-Level Semantic Prediction]
We extract high-level semantic representations of the input frames using a pretrained vision foundation model. These representations capture the essential structure of the scene while abstracting low-level details. In the first stage, the model predicts future frames in this semantic space, allowing it to focus on structural reasoning before synthesizing fine-grained visuals.    

\item[Semantics-Guided Video Generation]  
In the second stage, a latent video diffusion model generates future frames within the compact latent space of a Variational Autoencoder (VAE), which preserves sufficient detail for high-fidelity reconstruction. 
The previously predicted semantic representations guide the diffusion process, ensuring the generated content remains consistent with the scene’s semantic dynamics. Finally, the VAE decoder reconstructs the output frames in pixel space, producing photorealistic future frames aligned with the predicted semantics.
\end{description}  

An overview of our \ours framework during inference is shown in \autoref{fig:overview}.
This two-stage approach effectively separates semantic reasoning from visual synthesis, simplifying the video prediction task. \hyperref[sec:high_level_semantic]{Subsection~\ref*{sec:high_level_semantic}} 
details the semantic prediction stage, \hyperref[sec:sem_guided_generation]{Subsection~\ref*{sec:sem_guided_generation}} describes the semantics-guided generation process, \hyperref[sec:diffusion_network]{Subsection~\ref*{sec:diffusion_network}} presents the architecture of the diffusion model, and 
\hyperref[sec:training_strategies]{Subsection~\ref*{sec:training_strategies}} outlines the training strategies that ensure robust semantic conditioning.

\subsection{High-Level Semantic Prediction}
\label{sec:high_level_semantic}

In the first stage, we extract high-level semantic representations from the input frames using a pretrained Vision Foundation Model (VFM). Specifically, we employ the DINOv2 \cite{oquab2024dinov2} image encoder \( E_h(\cdot) \), which processes each frame \( x_t \) independently to produce a feature map \( h_t \):  
\begin{equation}
    h_t = E_h(x_t), \quad t = 1, \ldots, M
\end{equation}
Here, \( h_t \) has dimensions \( H_h \times W_h \times C_h \), where \( H_h \times W_h \) are the spatial dimensions and \( C_h \) is the number of feature channels. These features capture the scene’s semantic structure while abstracting low-level details.  

Next, we predict future features autoregressively using a feature generation model \( G_h(\cdot) \). Given the context features \( (h_1, \ldots, h_M) \), \( G_h(\cdot) \) generates the remaining \( K-M \) frames one step at a time:  
\begin{equation}
    h_{M+1}, \ldots, h_K = G_h(h_1, \ldots, h_M).
\end{equation}
We adopt the masked transformer architecture from \cite{karypidis2024dino} for \( G_h(\cdot) \). During training, the model takes \( M+1 \) frame features as input. The first \( M \) frames (context) are unmasked, while the features of the \( (M+1) \)-th frame (the prediction target) are entirely masked. The model is trained to regress the masked features using a Smooth L1 loss:  
\begin{equation}
    \mathcal{L}_{\text{feat}} = \text{SmoothL1}\left( G_h(h_1, \ldots, h_M), h_{M+1} \right).
\end{equation}

At inference time, the model operates autoregressively: given \( M \) context frames, it predicts the next frame’s features, which are then fed back as input for subsequent predictions. 
This stage ensures consistent semantic reasoning before proceeding to fine-grained synthesis in the next stage.

\subsection{Semantics-Guided Video Generation}
\label{sec:sem_guided_generation}

In the second stage, we use a latent video diffusion model to generate actual future frames, guided by the predicted semantic features. The model takes as input the context frames \( (x_1, \ldots, x_M) \), their semantic features \( (h_1, \ldots, h_M) \), and the predicted future semantic features \( (h_{M+1}, \ldots, h_K) \) from \( G_h(\cdot) \).  

First, the context frames are encoded into compact latent features using a causal 3D VAE encoder \( E_z(\cdot) \):  
\begin{equation}  
z_{1:M} = E_z(x_{1:M}). 
\end{equation}  
Here, \( z_t \) has dimensions \( H_z \times W_z \times C_z \), where \( H_z \times W_z \) are the spatial dimensions and \( C_z \) is the number of channels. 
For the 3D VAE, we employ the WAN2.1 VAE~\cite{wan2025wan}, a causal variational autoencoder that compresses videos along both spatial and temporal dimensions. For clarity, we omit the details of temporal subsampling. To align the temporal resolution between the two stages, the semantic prediction stage processes only every \( \tfrac{1}{r} \) frame, where \( r \) is the temporal subsampling ratio of the VAE encoder.

Next, we generate the future latent frames using a diffusion model \( G_z(\cdot) \). Following standard diffusion terminology, we gradually denoise latent features for the future frames. Let \( z_t^{(n)} \) denote the noisy latent of frame \( t \) at diffusion step \( n \). The forward process gradually adds Gaussian noise to the ground-truth future latents \( z_t \) (for \( t = M+1, \ldots, K \)):  
\begin{equation}  
z_t^{(n)} = z_t + \sigma_n \epsilon, \quad \epsilon \sim \mathcal{N}(0, I)  
\end{equation}
where \( \sigma_n \) controls the noise schedule.

The denoising model \( G_z(\cdot) \) takes as input the noised future latents \(z_{M+1:K}^{(n)} = (z_{M+1}^{(n)}, \ldots, z_K^{(n)}) \), the clean context latents \(z_{1:M} = (z_1, \ldots, z_M) \) (no noise added), all semantic features \(h_{1:K} = (h_1, \ldots, h_K) \) (no noise added), and the noise step \( n \). It predicts the clean future-frame latents \(z_{M+1:K}^{(0)} = z_{M+1:K}\) for the future frames:  
\begin{equation}
\hat{z}_{M+1:K} = G_z \big( z_{M+1:K}^{(n)};\, z_{1:M},\, h_{1:K},\, n\big).
\end{equation}
The model is trained to minimize the denoising objective: 
\begin{equation}  
\mathcal{L}_{\text{diffusion}} = \mathbb{E}_{n, \epsilon} \left[\lambda_{n} \left\| \hat{z}_{M+1:K} - z_{M+1:K} \right\|^2 \right]  
\end{equation}
where \(\lambda_{n}\) is a reweighting function that balances the contribution of different noise levels.  
Only the future frames (\( t = M+1, \ldots, K \)) contribute to this loss. 

Finally, the 3D VAE decoder \( D_z(\cdot) \) reconstructs the pixel-space frames from the latent sequence: 
\begin{equation}  
\hat{x}_{1:K} = D_z(\hat{z}_{M+1:K}).
\end{equation}

\subsection{\textbf{\ours} Architecture} \label{sec:diffusion_network}

\begin{wrapfigure}[37]{r}{0.5\linewidth}
\vspace{-17pt}
\centering
\includegraphics[trim={0cm 0cm 0cm 0cm},width=\linewidth]{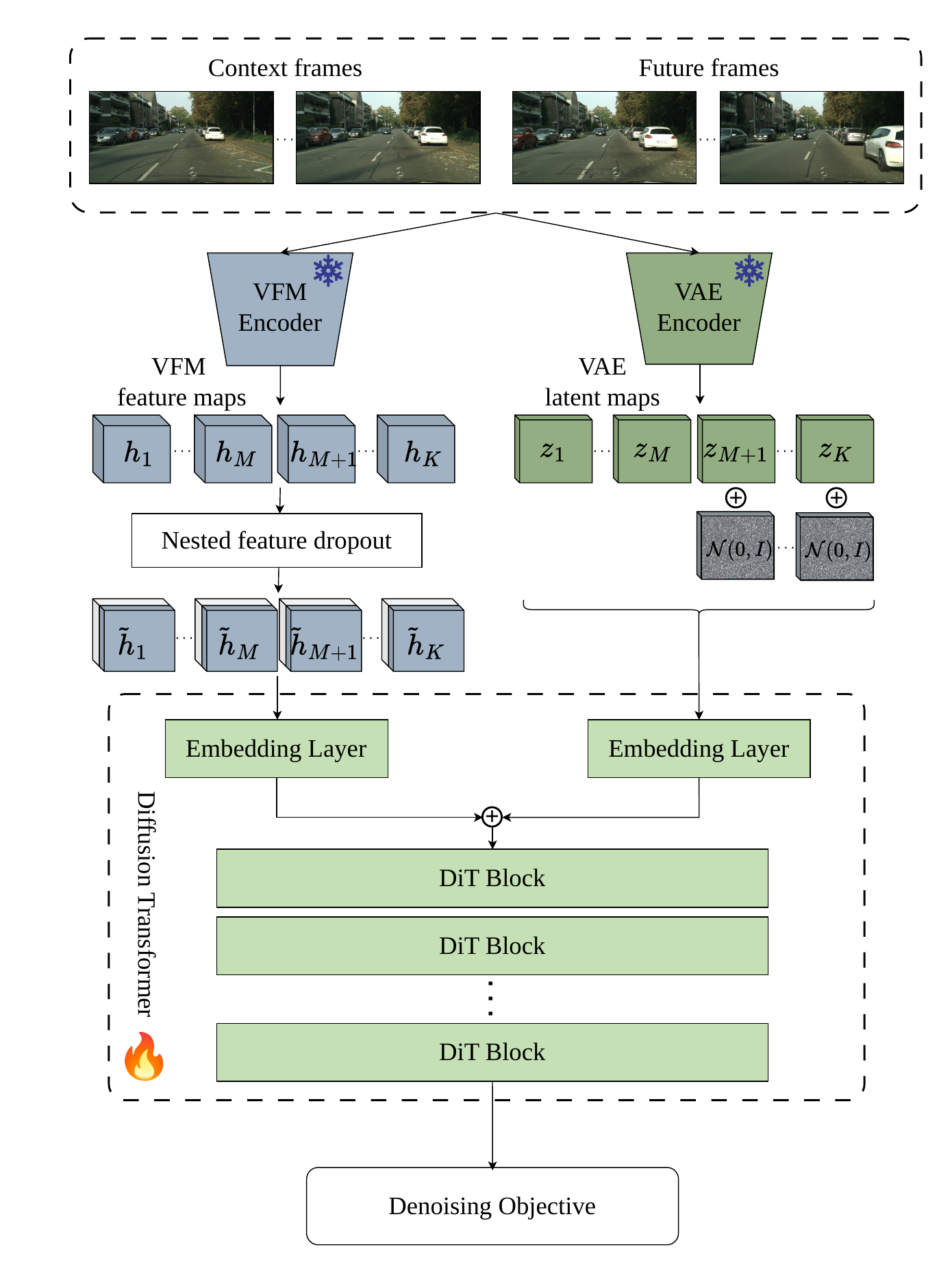}
\caption{\textbf{\ours architecture.} The model takes as input (1) VAE latents \( z_{1:K} \) with noise applied to the future frames \( z_{M+1:K} \), and (2) semantic features \( h_{1:K} \) from a vision foundation model (VFM), with nested dropout applied to zero-out fine-grained channels. The two modalities are embedded independently and combined via channel-wise summation at the input, implementing early fusion. The diffusion transformer is trained with a denoising objective applied to the future VAE latents. Nested dropout and mixed supervision (using for the future frames both ground-truth and predicted features with a 90\%/10\% mixture; not shown in the figure for simplicity) regularize the model against overfitting to idealized VFM features for the future frames.}
\label{fig:overview_training}
\end{wrapfigure}

The detailed architecture of the diffusion transformer during training, including early semantic alignment and nested dropout (discussed in
\hyperref[sec:training_strategies]{Subsection~\ref*{sec:training_strategies}}, is shown in \autoref{fig:overview_training}.

\paragraph{Diffusion architecture.}
The diffusion model \( G_z(\cdot) \) follows the video prediction design of Cosmos-Predict~\cite{agarwal2025cosmos,ali2025world}, which is built upon the Diffusion Transformer (DiT) framework~\cite{peebles2023scalable}. We retain the core architectural components of Cosmos-Predict, including 3D-factorized Rotary Position Embeddings (RoPE)~\cite{su2024roformer}, query/key normalization before attention~\cite{wortsman2023small, esser2024scaling, dehghani2023scaling}, and RMSNorm~\cite{zhang2019root} with learnable scales in all self-attention blocks.  
Noise-level conditioning is implemented via LoRA-based adaptive normalization (AdaLN-LoRA)~\cite{hu2022lora}, which replaces the parameter-heavy AdaLN layers of DiT, yielding a more efficient yet equally expressive design.

Compared to the original Cosmos-Predict architecture, we remove the cross-attention layers—since our model is not text-conditioned—and introduce a dedicated \emph{semantic guidance mechanism}, described next.

\paragraph{Early Semantic Alignment}
To incorporate semantic guidance, we fuse semantic features \( h \) with the VAE latent representations \( z \) at the input level. 
Specifically, the VAE latents \( z \) are patchified with a spatial size of \( 2 \times 2 \), and the semantic features \( h \) are resized to match the same spatial resolution. Both feature maps are then embedded independently and combined by channel-wise summation, ensuring joint conditioning from the outset. 
This design enables early semantic alignment between the predicted scene structure and the generative latent space, providing simple and efficient guidance for the diffusion process without increasing the token count.

\subsection{Training Strategies for Robust Semantic Conditioning}
\label{sec:training_strategies}

During training, we follow a teacher-forcing approach: the semantic features \( h_t \) for future frames are extracted from the ground-truth frames using the encoder \( E_h(\cdot) \).
While this setup stabilizes training and accelerates convergence, it introduces a mismatch between training and inference. At test time, the semantic features are provided by \( G_h \), which are inherently noisier than those extracted from ground truth. As a result, models trained exclusively with ground-truth features tend to overfit to these ideal representations—producing good semantic layouts but blurry pixel-level details, reflected by degraded FVD and FID scores.  

To mitigate this issue, we introduce two complementary strategies that improve robustness to imperfect semantic inputs.

\paragraph{Nested dropout of semantic input features.}
The semantic features \( h_t \in \mathbb{R}^{C_h \times H \times W} \) are derived from DINOv2~\cite{oquab2024dinov2} with a ViT-B backbone, by concatenating representations from the 3rd, 6th, 9th, and 12th transformer blocks and projecting them to $C_h=1152$ channels via PCA. These channels form a hierarchical representation, where higher-variance components capture coarse semantics and lower-variance components encode fine details.  

To prevent the diffusion model from over-relying on fine-grained features, we apply nested dropout~\cite{rippel2014learning, kusupati2022matryoshka} to all semantic feature maps \( h_{1:K} \).  
With equal probability, we retain only the first \(c \in {8, 16, 32, 64, 128, 256, 512, 1152} \) channels of each \( h_t \), replacing the remaining channels with zeros:  
\begin{equation} \label{eq:nested_dropout}  
\tilde{h}_{1:K} = \text{NestedDropout}(h_{1:K}; c) = [h_{1:K}^{1:c},\, \mathbf{0}^{C_h - c}],  
\end{equation}  
where \( \tilde{h}_{1:K} \) denotes the semantic features after nested dropout, \( h_{1:K}^{1:c} \) indicates the first \( c \) feature channels retained for all frames, and \([\,\cdot,\,\cdot\,]\) denotes the concatenation operation.
This stochastic truncation encourages the diffusion model to learn robust conditioning from the most informative semantic subspaces, rather than memorizing fine-scale correlations.

\paragraph{Mixed supervision with ground-truth and predicted semantic features.}
To further reduce the train–test gap, we train the diffusion model with a mixture of ground-truth and predicted semantic features.  
For each batch, we randomly sample 10\% of examples where the semantic features are taken from the generator \( G_h \) (predicted features), and 90\% where they come from \( E_h \) (ground-truth features).  
This mixture regularization exposes the diffusion model to both ideal and imperfect semantic inputs during training.  
Empirically, the 90/10 ratio provides the best trade-off—reducing blur and improving visual quality without compromising semantic consistency.

\section{Experiments}
\label{sec:experiments}

Our experiments assess how well the proposed hierarchical \ours framework predicts future video frames that are both semantically faithful and photorealistic. We evaluate on 
multiple datasets and compare against strong baselines, analyze training dynamics (in \hyperref[sec:comparison_results]{Subsection~\ref*{sec:comparison_results}}), and conduct detailed ablations (in  \hyperref[sec:ablations]{Subsection~\ref*{sec:ablations}}). Across metrics, our method yields consistent gains in temporal semantic consistency, generation quality, and training efficiency, 
demonstrating the effectiveness of coarse-to-fine hierarchical visual modeling.

\subsection{Experimental Setup}

\paragraph{Datasets.}
We experiment on four driving datasets: Cityscapes~\cite{Cordts_2016_CVPR},
nuScenes~\cite{nuscenes}, CoVLA~\cite{covla}, and KITTI~\cite{geiger2013vision}.
Our primary setup trains and evaluates on Cityscapes. To assess
generalization at larger scale, we additionally train on a combination
of Cityscapes, nuScenes, and CoVLA, evaluating in-domain on Cityscapes
and nuScenes, and zero-shot on KITTI. Full dataset details are provided in  \hyperref[sec:datasetdetails]{Appendix Section~\ref*{sec:datasetdetails}}.

\paragraph{Implementation Details.}
We use DINOv2-Reg ViT-B/14~\cite{oquab2024dinov2} as the VFM encoder for semantic feature extraction. Our diffusion transformer follows the EDM formulation~\cite{karras2022elucidating} used in Cosmos-Predict~\cite{agarwal2025cosmos}, with 14 layers, 16 attention heads, and a 2048-dimensional embedding, totaling roughly 800M parameters. 

We train on sequences of $K=25$ frames, where $M=13$ context frames at $432\times768$ resolution are encoded using the WAN2.1 VAE~\cite{wan2025wan} with $8\times8\times4$ temporal-spatial downsampling, yielding 7 latent frames of size $54\times96$. Models are trained from scratch using Adam~\cite{adamopt} ($\beta_1\!=\!0.9$, $\beta_2\!=\!0.99$) with a learning rate of $0.6 \times 2^{-10.5}$ and linear warmup and decay. Training runs for 40k iterations on 8 NVIDIA H200 GPUs for single dataset experiments and 120k iterations for multiple dataset experiments) with effective batch size 8, requiring approximately 7 and 28 hours respectively. Full implementation details regarding diffusion formulation, feature prediction model and semantic decoders are provided in \hyperref[sec:FullImplDetails]{Appendix Section~\ref*{sec:FullImplDetails}}.

\paragraph{Evaluation Protocol and Metrics.}
For all experiments, we use frames 3--15 as context and predict frames 16--27. We evaluate two complementary aspects:
\begin{itemize}
    \item \textbf{Temporal semantic consistency:} How well the generated future frames preserve the evolving scene semantics.
    \item \textbf{Generation quality:} How realistic and temporally coherent the synthesized frames appear.
\end{itemize}

\paragraph{Semantic consistency metrics.}
We evaluate semantic segmentation (mIoU) and depth estimation on the generated frame 19. Segmentation is computed using a DINOv2-Reg ViT-B encoder with DPT heads~\cite{ranftl2021vision}. We report mIoU over all classes (A) and over moving-object classes (M), which are more challenging and critical for autonomous driving scenarios. Depth evaluation uses Absolute Relative Error (AbsRel) and threshold accuracy ($\delta_1$), where lower AbsRel and higher $\delta_1$ indicate better performance. Following standard practice, frame 19 in each sequence provides dense annotations for 19 semantic classes. Since Cityscapes and nuScenes not include dense depth labels, we generate pseudo-depth using Depth Anything V2~\cite{yang2024depth}. 

\paragraph{Generation quality metrics.}
We compute FID~\cite{NIPS2017_fid} and FVD~\cite{unterthiner2018_fvd} over all predicted frames to capture both spatial realism and temporal coherence. Lower scores indicate better generative quality.

\subsection{Video Prediction Results} \label{sec:comparison_results}

\paragraph{Comparison with Baselines}
We compare our hierarchical approach with three baselines:
(i) a standard diffusion video model trained with a denoising objective,  
(ii) REPA~\cite{yu2024representation}, and  
(iii) VideoREPA~\cite{zhang2025videorepa}.  
For reference, we additionally report the semantic consistency results of the VFM feature forecasting model used at stage 1 as \ours (Stage 1).

\begin{table*}[t]
\caption{\textbf{Comparison with baselines.}  
All numbers report absolute performance; values in parentheses indicate improvement over the Baseline.  
Blue indicates improvement, red indicates degradation.  
Our hierarchical approach provides consistent gains across both semantic consistency and generation quality.  
Model parameters: Baseline (782M), REPA variants (792M), Baseline-Large (1.5B), \ours (1.1B).}
\footnotesize
\centering
\resizebox{\columnwidth}{!}{%
\begin{tabular}{l cccc cc}
\toprule
\mr{2}{\Th{Method}} & \multicolumn{4}{c}{\Th{Semantic Consistency}} & \multicolumn{2}{c}{\Th{Generation}} \\
\cmidrule(lr){2-5} \cmidrule(lr){6-7}
& mIoU(A)\Th{$\uparrow$} & IoU(M)\Th{$\uparrow$} & $\delta_1$\Th{$\uparrow$} & AbsR\Th{$\downarrow$} & \Th{FID}\Th{$\downarrow$} & \Th{FVD}\Th{$\downarrow$} \\
\midrule
\graycell{\ours (Stage-1)} & \graycell{69.76} & \graycell{69.66} & \graycell{88.03} & \graycell{.1262} & \graycell{-} & \graycell{-} \\
\midrule
Baseline & 60.55 & 57.64 & 85.15 & .1460 & 12.86 & 60.70 \\
w/ REPA \cite{leng2025repa} 
& 61.45 {\color{blue}(+0.90)} 
& 59.63 {\color{blue}(+1.99)} 
& 85.35 {\color{blue}(+0.20)} 
& .1465 {\color{red}(+0.0005)} 
& 12.34 {\color{blue}(-0.52)} 
& 55.15 {\color{blue}(-5.55)} \\

w/ VideoREPA \cite{zhang2025videorepa} 
& 60.98 {\color{blue}(+0.43)} 
& 58.61 {\color{blue}(+0.97)} 
& 85.22 {\color{blue}(+0.07)}
& .1466 {\color{red}(+0.0006)} 
& 12.95 {\color{red}(+0.09)} 
& 59.03 {\color{blue}(-1.67)} \\

Baseline-Large 
& 61.69 {\color{blue}(+1.14)} 
& 59.65 {\color{blue}(+2.01)} 
& 85.43 {\color{blue}(+0.28)} 
& .1453 {\color{red}(-0.0070)} 
& 11.99 {\color{blue}(-0.87)} 
& 56.81 {\color{blue}(-3.89)} \\

\ours 
& \textbf{63.53} {\color{blue}(\textbf{+2.98})} 
& \textbf{62.29} {\color{blue}(\textbf{+4.65})} 
& \textbf{85.72} {\color{blue}(\textbf{+0.57})} 
& \textbf{.1413} {\color{blue}(\textbf{-0.0047})} 
& \textbf{\,9.90} {\color{blue}(\textbf{-2.96})} 
& \textbf{52.66} {\color{blue}(\textbf{-8.04})} \\

\bottomrule
\end{tabular}
}
\label{tab:methods_comparison}
\vspace{-10pt}
\end{table*}

As shown in \autoref{tab:methods_comparison}, our \ours method achieves substantial improvements across both semantic consistency and generation quality. Segmentation and depth metrics show that \ours produces future frames with significantly improved temporal semantic fidelity. At the same time, \ours achieves the best FID and FVD scores, demonstrating its ability to synthesize photorealistic and temporally coherent predictions. These results confirm that jointly modeling semantic structure and pixel-space generation through a hierarchical design provides strong performance gains over existing methods. To account for the stochastic nature of our diffusion model, we additionally report mean and standard deviation over 3 independent sampling runs in \hyperref[subsec:sampling]{Appendix Subsection~\ref*{subsec:sampling}}.

\paragraph{Does \ours simply benefit from more parameters?}
To answer this question, we train a stronger baseline with additional diffusion transformer layers, resulting in a parameter count exceeding that of \ours. As shown in \autoref{tab:methods_comparison}, increasing the parameter count improves baseline performance but remains consistently inferior to \ours on both semantic and generation metrics. This indicates that the improvements arise from the hierarchical strategy rather than model size.

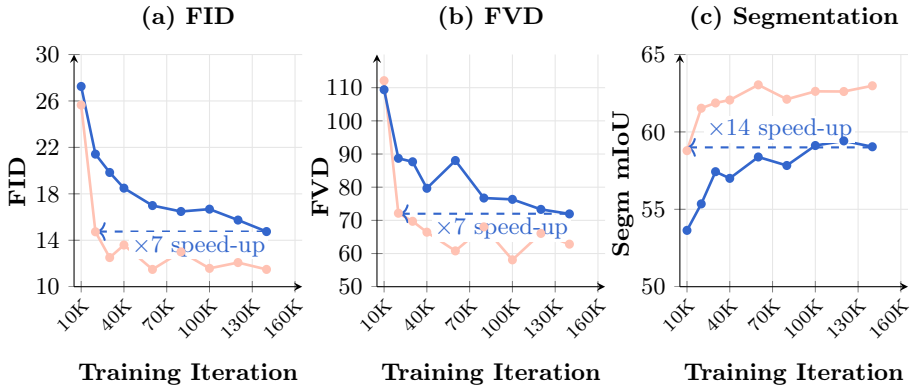
\begin{figure*}[t]
\centering
\footnotesize
\begin{tikzpicture}
    \begin{axis}[
      name=plot1,
      scale only axis,
      width=0.25 \textwidth,  
      height=0.25\textwidth,
      xmin=5000, xmax=165000,
      ymin=10.0, ymax=30.,
      xtick={10000, 40000, 70000, 100000, 130000, 160000},
      xticklabels={10K, 40K, 70K, 100K, 130K, 160K},
      scaled x ticks=false,
      xticklabel style={rotate=45, anchor=north east, font=\scriptsize, inner sep=1pt,},
      xlabel style={yshift=-4pt},  
      ytick={10, 14, 18, 22, 26, 30},
      xlabel={\textbf{Training Iteration}},
      ylabel={\textbf{FID}},
      title={\textbf{(a) FID}},
      y label style={at={(axis description cs:-0.25,0.6)}, anchor=east},
      grid=major, major grid style={gray!20},
      clip=false,
      axis lines=left,
    ]
      \addplot[color=softcoral!60, mark=*, line width=1.pt, mark size=1.2pt] 
        coordinates {
           (10000, 25.65) 
           (20000, 14.73) 
           (30000, 12.51) 
           (40000, 13.59)
           (60000, 11.49) 
           (80000, 13.00)
           (100000, 11.58) 
           (120000, 12.08)
           (140000, 11.49)
        };
      \addplot[color=softblue, mark=*, line width=1.pt, mark size=1.2pt] 
        coordinates {
           (10000, 27.25)
           (20000, 21.42) 
           (30000, 19.84) 
           (40000, 18.49)
           (60000, 16.99)
           (80000, 16.48)
           (100000, 16.68)
           (120000, 15.74)
           (140000, 14.75)
           
        };
      \draw[<-, dashed, softblue, line width=0.8pt]
        (axis cs:21000, 14.75) -- (axis cs:140000, 14.75)
        node[pos=0.6, below=0mm, fill=white, inner sep=1pt] {\(\times 7\) speed-up};
    \end{axis}

    \begin{axis}[
      name=plot2,
      at={($(plot1.east)+(1cm,0)$)}, anchor=west,  
      scale only axis,
      width=0.25\textwidth,
      height=0.25\textwidth,
      xmin=5000, xmax=165000,
      ymin=50, ymax=120,
      xtick={10000, 40000, 70000, 100000, 130000, 160000},
      xticklabels={10K, 40K, 70K, 100K, 130K, 160K},
      xticklabel style={rotate=45, anchor=north east, font=\scriptsize, inner sep=1pt,},
      xlabel style={yshift=-4pt},  
      scaled x ticks=false,
      ytick={50, 60, 70, 80, 90, 100, 110},
      xlabel={\textbf{Training Iteration}},
      ylabel={\textbf{FVD}},
      title={\textbf{(b) FVD}},
      y label style={at={(axis description cs:-0.25,0.6)}, anchor=east},
      grid=major, major grid style={gray!20},
      clip=false,
      axis lines=left,
    ]
      \addplot[color=softcoral!60, mark=*, line width=1.pt, mark size=1.2pt] 
        coordinates {
           (10000, 112.11) 
           (20000, 72.15) 
           (30000, 69.72) 
           (40000, 66.44)
           (60000, 60.80) 
           (80000, 68.04)
           (100000, 58.10)
           (120000, 66.07)
           (140000, 62.83) 
        };
      \addplot[color=softblue, mark=*, line width=1.pt, mark size=1.2pt] 
        coordinates {
           (10000, 109.39) 
           (20000, 88.71) 
           (30000, 87.66) 
           (40000, 79.67)
           (60000, 88.04) 
           (80000, 76.73)
           (100000, 76.33) 
           (120000, 73.31)
           (140000, 71.95) 
        };
      \draw[<-, dashed, softblue, line width=0.8pt]
        (axis cs:21000, 72) -- (axis cs:140000, 72)
        node[pos=0.6, below=-0mm, fill=white, inner sep=1pt] {\(\times 7\) speed-up};
    \end{axis}

    \begin{axis}[
      name=plot3,
      at={($(plot2.east)+(1cm,0)$)}, anchor=west,  
      scale only axis,
      width=0.25\textwidth,
      height=0.25\textwidth,
      xmin=5000, xmax=165000,
      ymin=50, ymax=65,
      xtick={10000, 40000, 70000, 100000, 130000, 160000},
      xticklabels={10K, 40K, 70K, 100K, 130K, 160K},
      scaled x ticks=false,
      xticklabel style={rotate=45, anchor=north east, font=\scriptsize, inner sep=1pt,},
      xlabel style={yshift=-4pt},  
      ytick={50, 55, 60, 65},
      yticklabels={50, 55, 60, 65},
      xlabel={\textbf{Training Iteration}},
      ylabel={\textbf{Segm mIoU}},
      title={\textbf{(c) Segmentation}},
      y label style={at={(axis description cs:-0.25, 0.8)}, anchor=east},
      grid=major, major grid style={gray!20},
      clip=false,
      axis lines=left,
    ]
      \addplot[color=softcoral!60, mark=*,line width=1.pt, mark size=1.2pt] 
        coordinates {
           (10000, 58.79) 
           (20000, 61.53) 
           (30000, 61.87) 
           (40000, 62.06)
           (60000, 63.04) 
           (80000, 62.11) 
           (100000, 62.62)
           (120000, 62.61)
           (140000, 62.98)
           
        };
      \addplot[color=softblue, mark=*,line width=1.pt, mark size=1.2pt] 
        coordinates {
           (10000, 53.64) 
           (20000, 55.35) 
           (30000, 57.43) 
           (40000, 57.00)
           (60000, 58.38)
           (80000, 57.83)
           (100000, 59.12)
           (120000, 59.42)
           (140000, 59.04)
        };
      \draw[<-, dashed, softblue, line width=0.8pt]
        (axis cs:10000, 59) -- (axis cs:140000, 59)
        node[pos=0.5, above=0.5mm, fill=white, inner sep=1pt] {\(\times 14\) speed-up};
    \end{axis}

\end{tikzpicture}
\caption{\textbf{Accelerated Training Convergence.} Training curves for (a) FID, (b) FVD, and (c) Segmentation mIoU comparing our hierarchical approach (orange) with the baseline (blue). Our method achieves $\times$7 speed-up for generation metrics and $\times$14 speed-up for segmentation.}
\label{fig:accelerated_convergence}
\vspace{-15pt}
\end{figure*}

\paragraph{\ours accelerates diffusion training.}
Recent representation-alignment methods for images~\cite{yu2024representation} have shown that semantically aligned guidance can improve convergence. We test whether our hierarchical approach offers similar benefits in video generation.
To study convergence dynamics, we train both the baseline and \ours for 140k iterations using a constant learning rate after warmup, enabling direct comparison across checkpoints. As shown in \autoref{fig:accelerated_convergence}, \ours outperforms the baseline throughout training for all metrics:
\begin{itemize}
    \item For FID, \ours reaches 15 in \textbf{20k iterations}, whereas the baseline requires \textbf{140k}—a \textbf{$7\times$ speed-up}. FVD exhibits a similar \textbf{$7\times$ acceleration}.  
    \item Semantic metrics also converge faster: segmentation mIoU achieves a \textbf{$14\times$ speed-up}.
\end{itemize}
Qualitative comparisons across diverse driving scenes are provided in \hyperref[sec:qualitative]{Appendix Section~\ref*{sec:qualitative}}.

\paragraph{Scaling and Cross-Dataset Generalization.} To evaluate whether the benefits of our hierarchical framework scale with data diversity, we train \ours on a larger combined dataset (Cityscapes, nuScenes, and CoVLA) and assess performance across both in-domain (Cityscapes and nuScenes) and zero-shot (KITTI) settings. As shown in Table~\ref{tab:scale_combined}, our method consistently outperforms the baselines (Baseline and Baseline Large) across all benchmarks: on Cityscapes and nuScenes, it improves both semantic consistency and generation quality, while on KITTI, it demonstrates superior zero-shot generalization.
We further compare our approach against state-of-the-art large-scale systems, specifically Vista\cite{gao2024vista} and Cosmos-Predict 2\cite{cosmos_predict2}. Although these models are pretrained with several orders of magnitude more data and compute—and were subsequently fine-tuned in our specific setting—\ours surpasses Vista and performs competitively with Cosmos-Predict 2. This result is particularly significant as it demonstrates that our hierarchical design achieves comparable—and in some cases superior—results with a fraction of the training overhead.

These results demonstrate that incorporating hierarchical semantic guidance not only improves final performance but also substantially accelerates training.

\subsection{Ablation Study} \label{sec:ablations}

\begin{table*}[t!]
\caption{\emph{Results for \ours trained on extended data (Cityscapes + nuScenes +
CoVLA). We report in-domain results on Cityscapes and nuScenes, zero-shot
generalization on KITTI, and comparisons against \ours~(Stage~1), Baselines and
large-scale internet-pretrained systems (Vista, Cosmos-Predict-2), both fine-tuned
in our setting. For reference, we include \ours (Stage~1), which reports semantic consistency metrics only, as it operates purely in feature space without the ability of generating pixels.
}}
\vspace{-8pt}
\centering
\setlength{\tabcolsep}{4pt}
\resizebox{\textwidth}{!}{%
\begin{tabular}{l cccc cc cc cc cc}
\toprule
\mr{3}{\Th{Method}} &
\multicolumn{6}{c}{\Th{Cityscapes}} &
\multicolumn{4}{c}{\Th{nuScenes}} &
\multicolumn{2}{c}{\Th{KITTI}} \\
\cmidrule(lr){2-7} \cmidrule(lr){8-11} \cmidrule(lr){12-13}
&
\multicolumn{4}{c}{\Th{Semantic Consistency}} &
\multicolumn{2}{c}{\Th{Generation}} &
\multicolumn{2}{c}{\Th{Sem. Cons.}} &
\multicolumn{2}{c}{\Th{Generation}} &
\multicolumn{2}{c}{\Th{Generation}} \\
\cmidrule(lr){2-5} \cmidrule(lr){6-7} \cmidrule(lr){8-9} \cmidrule(lr){10-11} \cmidrule(lr){12-13}
& \Th{mIoU(A)}$\uparrow$ & \Th{IoU(M)}$\uparrow$ & $\delta_{1}$$\uparrow$ & \Th{AbsR}$\downarrow$
& \Th{FID}$\downarrow$ & \Th{FVD}$\downarrow$
& $\delta_{1}$$\uparrow$ & \Th{AbsR}$\downarrow$
& \Th{FID}$\downarrow$ & \Th{FVD}$\downarrow$
& \Th{FID}$\downarrow$ & \Th{FVD}$\downarrow$ \\
\midrule
\addlinespace[2pt]
\multicolumn{13}{l}{\graycell{\emph{VFM forecasting only}}} \\
\addlinespace[2pt]
\,\,\graycell{Re2Pix (Stage 1)} & \graycell{70.46} & \graycell{70.26} & \graycell{88.03} & \graycell{.1208} & \graycell{-} & \graycell{-} & \graycell{83.82} & \graycell{.1869} & \graycell{-} & \graycell{-} & \graycell{-} & \graycell{-} \\
\midrule
\addlinespace[2pt]
\multicolumn{13}{l}{\emph{Trained from scratch}} \\
\addlinespace[2pt]
\,\,Baseline           & 62.77 & 60.86 & 85.62 & .1433 &  9.64 & 52.12 & 79.92 & .2438 & 21.08 & 136.71 & 20.02 & 61.02 \\
\,\,Baseline (Large)   & 63.32 & 61.73 & 85.63 & .1423 &  9.89 & 51.77 & 80.17 &  .2412     & 20.73 &  134.58 & 20.94 & 61.28     \\

\,\,\ours (ours)      & 64.63 & 63.52 & 86.01 & .1400 &  9.29 & 49.03 & 81.09 & .2306 & 18.96 & 134.03 & 15.94 & 61.73 \\
\midrule
\addlinespace[2pt]
\multicolumn{13}{l}{\graycell{\emph{Large-scale pretrained}}} \\
\addlinespace[2pt]
\,\,\graycell{Vista (Finetuned)}  & \graycell{63.88} & \graycell{62.35} & \graycell{85.91} & \graycell{.1376} & \graycell{12.17} & \graycell{84.14} & \graycell{80.46} & \graycell{.2304} & \graycell{21.06} & \graycell{174.86} & \graycell{7.95} & \graycell{97.96}  \\
\,\,\graycell{Cosmos-Predict-2 (Finetuned)} & \graycell{65.69} & \graycell{64.85} & \graycell{86.19} & \graycell{.1371} &  \graycell{7.74} & \graycell{46.22} & \graycell{81.28} & \graycell{.2243} & \graycell{15.90}  & \graycell{133.19} & \graycell{8.62}  &  \graycell{41.64}     \\
\bottomrule
\end{tabular}%
}
\vspace{-15pt}
\label{tab:scale_combined}
\end{table*}

\paragraph{Effect of Nested Dropout.}
We evaluate the impact of nested dropout by comparing two settings (\autoref{tab:ablation_nested}): using a fixed set of 1152 feature channels, and applying nested dropout during training (with the full 1152 at test time). Nested dropout yields consistent improvements across all metrics. The largest gains appear in FID and FVD, indicating substantially better generation quality.  
By stochastically truncating fine-grained semantic channels during training, nested dropout encourages the diffusion model to rely on robust, coarse-to-fine semantic structure rather than overfitting to perfect ground-truth feature values. This reduces the train–test mismatch between ground-truth features (seen during training) and predicted features (used at inference), enabling sharper and more realistic synthesis without degrading semantic fidelity. Building on nested dropout, we further investigate a CFG-inspired representation guidance scheme that contrasts predictions at different levels of semantic granularity. Ablations on this technique property are provided \hyperref[subsec:repguid]{Appendix Subsection~\ref*{subsec:repguid}}.

\paragraph{Mixed supervision with ground-truth and predicted features.}
We compare three training configurations (Table~\ref{tab:ablation_training_features}):  
(i) conditioning on ground-truth VFM features,  
(ii) conditioning on predicted features, and  
(iii) our mixed strategy (90\% ground-truth, 10\% predicted).  

    Training only on ground-truth features yields the strongest semantic consistency, but severely degrades perceptual quality due to a train–test mismatch: the model overfits to precise ground-truth feature values for generating fine-grained image details and struggles when exposed to noisier predicted features at inference, producing blurry frames. Training only on predicted features reverses this trend—generation metrics improve, but perception metrics degrade.  

Our mixed strategy combines the advantages of both: it matches the perception quality of the ground-truth-only model while substantially improving FID and FVD. These results highlight that stochastically mixing the two sources enables robust semantic conditioning while preserving high-fidelity synthesis.

\begin{table*}[t]
\begin{minipage}[t]{0.44\linewidth}
    \caption{\textbf{Impact of Nested Dropout.} Training with nested dropout (bottom) substantially improves generation quality compared to training with fixed 1152 components (top), while also improving the semantic consistency metrics.}
    \resizebox{\linewidth}{!}{%
    
\footnotesize
\centering
\begin{tabular}{l cccc cc} \toprule
\mr{2}{\Th{Components}} & \multicolumn{4}{c }{\Th{Sem. Consistency}} & \multicolumn{2}{c}{\Th{Generation}} \\ 
\cmidrule(lr){2-5} \cmidrule(lr){6-7}
 & mIoU(A)\Th{$\uparrow$} & IoU(M)\Th{$\uparrow$} & $\delta_1$\Th{$\uparrow$} & AbsR\Th{$\downarrow$}  & \Th{FID}$\downarrow$  & \Th{FVD}$\downarrow$ \\
\midrule 
Fixed (1152) & 62.38 & 60.67 & 85.45 & .1467 & 12.63 & 58.91 \\ 
Nested & \textbf{63.53} & \textbf{62.29} & \textbf{85.72} & \textbf{.1413} & \textbf{9.90} & \textbf{52.66} \\ 
\bottomrule
\end{tabular}
\label{tab:ablation_nested}

    }
\end{minipage}
\hfill
\begin{minipage}[t]{0.52\linewidth}
    \caption{\textbf{Impact of mixed supervision with GT and predicted features.} Training with a mixture of ground-truth (90\%) and predicted features (10\%) combines the semantic consistency benefits of training with ground truth features with the generation quality benefits of training with predicted features.}
    \resizebox{\linewidth}{!}{%

\footnotesize
\centering
\setlength{\tabcolsep}{2.5pt}
\begin{tabular}{l cccc cc} \toprule
\mr{2}{\Th{Features}} & \multicolumn{4}{c}{\Th{Sem. Consistency}} & \multicolumn{2}{c}{\Th{Generation}} \\ 
\cmidrule(lr){2-5} \cmidrule(lr){6-7}
 & mIoU(A)\Th{$\uparrow$} & IoU(M)\Th{$\uparrow$} & $\delta_1$\Th{$\uparrow$} & AbsR\Th{$\downarrow$} & \Th{FID}$\downarrow$ & \Th{FVD}$\downarrow$ \\
\midrule 
Baseline & 60.55 & 57.64 & 85.15 & .1460 & 12.86 & 60.70  \\
\midrule
Ground Truth  & 64.42 & 63.15 & 86.06 & .1407 & 10.43 & 80.85 \\ 
Predicted     & 62.77 & 61.38 & 85.58 & .1420 & 10.21 & 55.81 \\
Mixed (90/10) & 63.53 & 62.29 & 85.72 & .1413 & 9.90 & 52.66 \\ 
\bottomrule
\end{tabular}
\label{tab:ablation_training_features}

    }
\end{minipage}
\end{table*}

\paragraph{Number of semantic components at inference.}
Because nested dropout exposes the model to varying semantic feature dimensionalities during training, we can adjust the number of PCA components at inference time. As shown in \autoref{tab:ablation_inference_feats}, performance remains strong even when reducing to 128 components, indicating that coarse semantic structure captured by top principal components is highly informative. Performance degrades gracefully at very low dimensions (e.g., 8 or 16 components), while the full 1152 components yield the best scores w.r.t. the semantic consistency metrics.

\paragraph{Sensitivity to VFM features.}
To assess whether \ours depends on a specific vision foundation model, we replace DINOv2 with SigLIP-2\cite{tschannen2025siglip2} as the feature extractor in Stage~1 and retrain on Cityscapes (Table~\ref{tab:ablation_vfm}). Both variants consistently outperform the Baseline across semantic consistency and generation metrics,
demonstrating that the hierarchical design is robust to the choice of VFM, with DINOv2 yielding slightly stronger results consistent.

\begin{table*}[t]
\begin{minipage}[t]{0.44\linewidth}
    \caption{\textbf{Number of semantic components at inference.} Performance on Cityscapes across different numbers of PCA components at inference. Using 256 components achieves comparable results to the full 1152 components.}
    \resizebox{\linewidth}{!}{%
    
\centering
\setlength{\tabcolsep}{2.5pt}
\begin{tabular}{r cccc cc} \toprule
\mr{2}{\Th{Components}} & \multicolumn{4}{c }{\Th{Sem. Consistency}} & \multicolumn{2}{c}{\Th{Generation}} \\ 
\cmidrule(lr){2-5} \cmidrule(lr){6-7}
 & mIoU(A)\Th{$\uparrow$} & IoU(M)\Th{$\uparrow$} & $\delta_1$\Th{$\uparrow$} & AbsR\Th{$\downarrow$}  & \Th{FID}$\downarrow$ & \Th{FVD}$\downarrow$ \\
\midrule 
8 & 62.07 & 60.33  & 85.52 & .1441 & 10.44 & 54.20 \\ 
16 & 62.56 & 60.88  & 85.51 & .1436 & 10.22 & 55.05 \\ 
32  & 63.09 & 61.81  & 85.59 & .1431 & 9.95 & 54.16 \\ 
64  & 63.19 & 61.75  & 85.67 & .1420 & 9.94 & 53.34 \\ 
128  & 63.27 & 61.72  & 85.56 & .1438 & 9.80 & 52.82 \\ 
256  & 63.44 & 62.10  & 85.80 & .1417 & 9.88 & 52.46 \\ 
512  & 63.46 & 62.22  & 85.67 & .1430 & 9.95 & 52.00 \\ 
1152  & 63.53 & 62.29 & 85.72 & .1413 & 9.90 & 52.66 \\
\bottomrule
\end{tabular}
\label{tab:ablation_inference_feats}

    }
\end{minipage}
\hfill
\begin{minipage}[t]{0.52\linewidth}
    \caption{\textbf{Sensitivity to VFM features.} We replace DINOv2 with SigLIP-2 as the feature extractor in Stage~1 and report results on Cityscapes. Both variants consistently outperform the Baseline, demonstrating that the hierarchical design is robust to the choice of VFM}
    \resizebox{\linewidth}{!}{%
    
\centering
\setlength{\tabcolsep}{4pt}
\begin{tabular}{l cccc cc}
\toprule
\mr{2}{\Th{Method}} &
\multicolumn{4}{c}{\Th{Semantic Consistency}} &
\multicolumn{2}{c}{\Th{Generation}} \\
\cmidrule(lr){2-5} \cmidrule(lr){6-7}
& \Th{mIoU(A)}$\uparrow$ & \Th{IoU(M)}$\uparrow$ & $\delta_{1}$$\uparrow$ & \Th{AbsR}$\downarrow$ & \Th{FID}$\downarrow$ & \Th{FVD}$\downarrow$ \\
\midrule
Baseline & 60.55 & 57.64 & 85.15 & .1460 & 12.86 & 60.70  \\
\ours w/ SigLIP2  &  63.02 & 61.51 &  85.73 & .1419 &  10.26   &  52.69   \\
\ours w/ DinoV2 & 63.53 & 62.29 & 85.72 & .1413 & 9.90 & 52.66 \\
\bottomrule
\end{tabular}%

\label{tab:ablation_vfm}

    }
\end{minipage}

\end{table*}

\section{Conclusion}
\label{sec:Conclusion}

We introduced \ours, a hierarchical semantic-to-pixel framework for video prediction that integrates VFM representation forecasting with a diffusion-based video generator. By first predicting future semantic representations and then leveraging them to guide pixel-space synthesis, \ours produces videos that are semantically consistent, temporally coherent, and photorealistic. Extensive experiments on multiple datasets demonstrate substantial improvements over strong baselines, including REPA, across temporal semantic consistency, perceptual quality, and training efficiency, with significant acceleration in convergence. Ablation studies further highlight the importance of nested dropout and mixed supervision for robust semantic conditioning and high-fidelity generation. Overall, our results show that explicitly modeling hierarchical semantic structure is an effective and scalable strategy for generating realistic future video frames, paving the way for more reliable video prediction in complex dynamic scenes.
\paragraph{Acknowledgements}

This work has been partially supported by project MIS 5154714 of the National Recovery and Resilience Plan Greece 2.0 funded by the European Union under the NextGenerationEU Program. Hardware resources were granted with the support of GRNET. Also, this work was performed using HPC resources from GENCI-IDRIS (Grants AD011016639 and AS011017163).



%
%
\bibliographystyle{splncs04}
\bibliography{main}
\clearpage

{\huge \textbf{Appendix}}

\section{Additional Results}

\subsection{Multiple Sampling for Stochastic Predictions}
\label{subsec:sampling}
Since our diffusion-based video prediction model produces stochastic outputs, we perform 3 independent sampling runs for each sequence and report the mean and standard deviation across samples to provide more robust performance estimates in Table~\ref{tab:sampling_results}. Averaging over multiple runs yields slightly better semantic metrics for both methods but modestly higher FVD. 
\begin{table*}[h]
\caption{\textbf{Multiple sampling results.} 
Single-run performance (top) and mean $\pm$ standard deviation over 3 independent sampling runs (bottom) for both methods.}\centering
\resizebox{\columnwidth}{!}{%
\begin{tabular}{l cccc cc}
\toprule
\mr{2}{\Th{Method}} & \multicolumn{4}{c}{\Th{Semantic Consistency}} & \multicolumn{2}{c}{\Th{Generation}} \\
\cmidrule(lr){2-5} \cmidrule(lr){6-7}
& mIoU(A)\Th{$\uparrow$} & IoU(M)\Th{$\uparrow$} & $\delta_1$\Th{$\uparrow$} & AbsR\Th{$\downarrow$} & \Th{FID}\Th{$\downarrow$} & \Th{FVD}\Th{$\downarrow$} \\
\midrule
Baseline (1 run) & 60.55 & 57.64 & 85.15 & .1460 & 12.86 & 60.70 \\
\ours (1 run) & 63.53 & 62.29 & 85.72 & .1413 & 9.90 & 52.66 \\
\midrule
Baseline (3 runs)
& 60.92 $\pm$ 0.34 
& 58.63 $\pm$ 0.86
& 85.25 $\pm$ 0.14 
& .1458 $\pm$ .0002
& 12.75 $\pm$ 0.10 
& 63.39 $\pm$ 2.63 \\

\ours (3 runs)
& 63.70 $\pm$ 0.23
& 62.76 $\pm$ 0.51 
& 85.75 $\pm$ 0.10 
& .1414 $\pm$ .0006 
& 9.91 $\pm$ 0.05 
& 55.16 $\pm$ 2.19 \\
\bottomrule
\end{tabular}
}
\label{tab:sampling_results}
\end{table*}

\subsection{CFG-style Representation Guidance with Nested Feature Dropout}
\label{subsec:repguid}
Leveraging the hierarchical ordering of semantic feature channels induced by the PCA projection, together with the nested dropout applied during training of the diffusion video generation model, we investigate a representation-guidance scheme inspired by classifier-free guidance (CFG) technique~\cite{karras2024guiding, ho2022classifier}. During sampling from the semantics-guided video diffusion model, we run two parallel forward passes at each diffusion step: one conditioned on all feature channels of $\tilde{h}_{1:K}$ ($C_h = 1152$ components), providing a high-fidelity prediction, and another conditioned on a nested subset $\tilde{h}_{1:K}^{1:c}$ with only $c < C_h$ components, yielding a coarser semantic signal via NestedDropout($\tilde{h}_{1:K}; c$). The final prediction is computed as:  
\begin{equation}  
\hat{z}_{M+1:K} = z_{M+1:K}^{(C_h)} + w \cdot (\hat{z}_{M+1:K}^{(C_h)} - \hat{z}_{M+1:K}^{(c)}),  
\end{equation}  
where $\hat{z}_{M+1:K}^{(c)}$ is the prediction using only the top-$c$ components, and $w$ controls the guidance strength. This formulation enhances fine-grained semantic details by contrasting predictions at different levels of semantic granularity.

\autoref{tab:cfg_ablation_c} ablates the choice of $c$ for a fixed guidance weight, while \autoref{tab:cfg_ablation_w} varies $w$ for a fixed $c=128$. The results show that this CFG-inspired representation guidance can selectively improve semantic fidelity or overall generation quality depending on the hyperparameter configuration. Although we do not enable this mechanism in our main experiments, it is an interesting emergent property of our approach and may further improve performance with additional exploration. We leave this for future work.

\begin{table*}[t]
\begin{minipage}[t]{0.50\linewidth}
    \caption{\textbf{Impact of component count $c$ in nested representation guidance.} Performance when using different numbers of nested PCA components in the CFG-style representation guidance formulation with $w=0.4$.}
    \resizebox{\linewidth}{!}{%
\footnotesize
\centering
\setlength{\tabcolsep}{3.5pt}
\begin{tabular}{r cccc cc} \toprule
\mr{2}{\Th{Components}} & \multicolumn{4}{c }{\Th{Sem. Consistency}} & \multicolumn{2}{c}{\Th{Generation}} \\ 
\cmidrule(lr){2-5} \cmidrule(lr){6-7}
 & mIoU(A)\Th{$\uparrow$} & IoU(M)\Th{$\uparrow$} & $\delta_1$\Th{$\uparrow$} & AbsR\Th{$\downarrow$}  & \Th{FID}$\downarrow$ & \Th{FVD}$\downarrow$ \\
\midrule 
8 & 62.73 & 61.09  & 85.49 & .1407 & 9.43 & 56.03 \\ 
16 & 63.39 & 62.22  & 85.64 & .1400 & 9.31 & 58.74 \\ 
32  & 63.73 & 62.60  & 85.74 & .1398 & 9.09 & 56.89 \\ 
64  & 64.23 & 63.45  & 85.76 & \textbf{.1398} & 9.08 & 57.25 \\ 
128 & \textbf{64.30} & \textbf{63.15} & 85.68 & .1399 & \textbf{8.97} & 55.32 \\
256  & 64.10 & 63.84  & \textbf{85.80} & .1399 & 9.10 & \textbf{54.43} \\ 
512  & 64.29 & 62.21  & 85.79 & .1403 & 9.10 & 54.88 \\ 
\bottomrule
\end{tabular}

\label{tab:cfg_ablation_c}
    }
\end{minipage}
\hfill
\begin{minipage}[t]{0.46\linewidth}
    \caption{\textbf{Impact of $w$ in nested representation guidance.} Performance when using different guidance weights $w$ in the CFG-style representation guidance formulation with number of components $c=128$.}
    \resizebox{\linewidth}{!}{%
\footnotesize
\centering
\setlength{\tabcolsep}{3.5pt}
\begin{tabular}{r cccc cc} \toprule
\mr{2}{\Th{w}} & \multicolumn{4}{c }{\Th{Sem. Consistency}} & \multicolumn{2}{c}{\Th{Generation}} \\ 
\cmidrule(lr){2-5} \cmidrule(lr){6-7}
 & mIoU(A)\Th{$\uparrow$} & IoU(M)\Th{$\uparrow$} & $\delta_1$\Th{$\uparrow$} & AbsR\Th{$\downarrow$}  & \Th{FID}$\downarrow$ & \Th{FVD}$\downarrow$ \\
\midrule 
0.9 & 64.39 & 63.33 & 85.63 & \textbf{.1378} & 9.53 & 66.38 \\ 
0.8 & 64.35 & 63.25 & 85.68 & .1385 & 9.34 & 63.30 \\ 
0.7 & 64.46 & \textbf{63.46} & 85.65 & .1391 & 9.23 & 61.60 \\ 
0.6 & 64.34 & 63.26 & 85.73 & .1390 & 9.07 & 59.54 \\ 
0.5 & \textbf{64.41} & 63.42  & 85.72 & .1395 & 9.01 & 57.39 \\ 
0.4 & 64.30 & 63.15  & 85.68 & .1399 & \textbf{8.97} & 55.32 \\ 
0.3 & 64.16 & 63.02  & \textbf{85.75} & .1404 & 9.03 & 53.74 \\
0.2 & 63.95 & 62.86  & 85.71 & .1409 & 9.14 & 52.89 \\ 
\midrule
0.0 & 63.53 & 62.29 & 85.72 & .1413 & 9.90 & \textbf{52.66} \\
\bottomrule
\end{tabular}

\label{tab:cfg_ablation_w}
    }
\end{minipage}
\end{table*}

\section{Qualitative Results}
\label{sec:qualitative}

\begin{figure*}[t!]
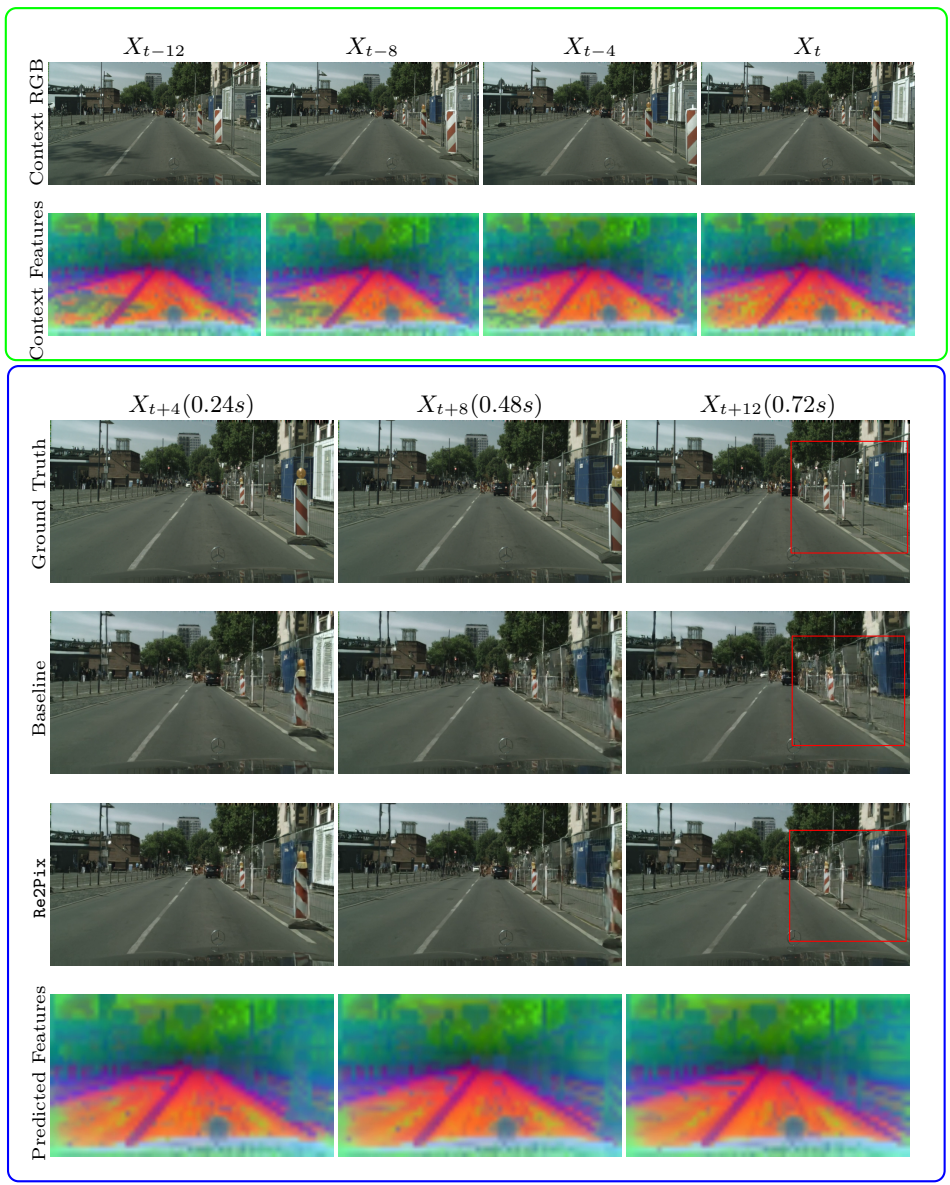

    \centering
    \include{figs/fig_supp_visual_60}
    \vspace{-20pt}
    \caption{\textbf{Qualitative example on Cityscapes (scene 60).} Top: context frames with corresponding DINOv2 features (PCA visualization). Bottom: ground-truth future frames, predictions from the baseline, and predictions from \ours with their forecasted semantic features. Red boxes highlight regions where \ours provides noticeably more accurate than the baseline.}
    \label{fig:qualitative-scene_60}
\end{figure*}

In \cref{fig:qualitative-scene_228,fig:qualitative-scene_60,fig:qualitative-scene_289,fig:qualitative-scene_456,fig:qualitative-scene_485} we present representative qualitative results comparing our method with the baseline. 
The context frames, marked with a green border and shown in the top rows, visualize both the input RGB sequences and the associated DINO pca features. Future frames (blue border) are plotted in the bottom section, each row showing (from top to bottom) the ground-truth, baseline predictions, \ours predictions, and the predicted semantic features that guide the generation of \ours. These visualizations allow for direct frame-by-frame comparison of temporal consistency, scene structure, and semantic fidelity between approaches.

Across the five diverse scenes presented, \ours more faithfully preserves scene geometry and object boundaries compared to the baseline. Our model also exhibits greater stability in open-road and urban traffic scenarios, maintaining sharper semantic structures and smoother future feature rollouts.

\section{Dataset Details}
\label{sec:datasetdetails}
We provide details on the four datasets used in our experiments.
\textbf{Cityscapes}~\cite{Cordts_2016_CVPR} contains video sequences captured from a vehicle driving through diverse urban environments across 50 cities. It comprises 2,975 training and 500 validation
sequences, each consisting of 30 frames recorded at 16\,fps at $1024\times2048$ pixel resolution.
\textbf{nuScenes}~\cite{nuscenes} is a large-scale dataset collected across Boston and Singapore, containing 1,000 driving scenes of 20 seconds each recorded at 12\,fps at $900\times1600$ pixel resolution, split into 750 training and 150 validation scenes.
\textbf{CoVLA}~\cite{covla} is a large-scale driving video dataset comprising 10,000 clips of approximately 30 seconds each, totaling over 80 hours of real-world driving footage recorded at 20\,fps at $1208\times1928$ pixel resolution, split into 8,000 training and 2,000 validation clips.
\textbf{KITTI}~\cite{geiger2013vision} contains driving sequences recorded in Karlsruhe at 10\,fps at $375\times1242$ pixel resolution, used solely for zero-shot generalization evaluation with no KITTI data seen during training. For all datasets, frames are resized to $432\times768$ pixels while preserving the original frame rate of each dataset.

\section{Implementation Details}
\label{sec:FullImplDetails}
\subsection{Diffusion Formulation Details}
Following the Cosmos-Predict \cite{agarwal2025cosmos} and EDM~\cite{karras2022elucidating} frameworks, we parameterize the noise level distribution using a log-normal distribution. For each training iteration, the noise level $\sigma_n$ is sampled as:
\begin{equation}
\sigma_n = \text{sigmoid}(u) = \frac{1}{1 + e^{-u}}, \quad u \sim \mathcal{N}(0,1),
\end{equation}
To ensure the model generalizes to the full noise range required during sampling, 5\% of training samples augment the log-normal distribution with very high noise levels sampled from $\log \sigma_n \sim \mathcal{U}(\log 200, \log 100000)$. This prevents the model from becoming biased towards low-noise regimes and improves denoising performance during the initial diffusion sampling steps.

Additionally, we apply input and output preconditioning \cite{karras2022elucidating} to stabilize training. The preconditioning coefficients are computed as:
\begin{align}
t_n &= \frac{\sigma_n}{1 + \sigma_n}, \\
c_{\text{skip}} &= 1 - t_n, \quad c_{\text{out}} = -t_n, \\
c_{\text{in}} &= 1 - t_n, \quad c_{\text{noise}} = t_n.
\end{align}
where $c_{\text{noise}}= 0.001$ for context frames.

The noise-level-dependent weighting $\lambda_n$ on the final diffusion loss: 
\begin{equation}  
\mathcal{L}_{\text{diffusion}} = \mathbb{E}_{n, \epsilon} \left[\lambda_{n} \left\| \hat{z}_{M+1:K} - z_{M+1:K} \right\|^2 \right]  
\end{equation}
is computed as:
\begin{equation}
\lambda_n = \frac{(1 + \sigma_n)^2}{\sigma_n^2}.
\end{equation}

\subsection{Feature Prediction Model}
Our feature prediction model adopts the DINO-Foresight \cite{karypidis2024dino} architecture, which uses a masked feature transformer to predict future VFM representations. By default, we use DINOv2-Reg with ViT-B/14 as the VFM visual encoder. To align the temporal resolution of Stage~1 with the WAN2.1
VAE temporal subsampling ratio of $r=4$, we extract DINOv2 features from every 4th frame of the input sequence, yielding one feature map per VAE latent frame. The masked feature transformer is built upon the \cite{besnier2023pytorch} implementation, consisting of 12 transformer layers with a hidden dimension of $d = 1152$ and sequence length $\mathcal{N} = 5$ (with $N_c = 4$ context frames and $N_p = 1$ future frame). For end-to-end training, we use the Adam optimizer \cite{adamopt} with momentum parameters $\beta_1 = 0.9$, $\beta_2 = 0.99$, and a learning rate of $6.4 \times 10^{-4}$ with cosine annealing. Training is conducted on 8 A100 40GB GPUs with an effective batch size of 64.

\subsection{Semantic Decoders}
We train DPT ~\cite{ranftl2021vision} decoder heads for semantic segmentation and depth estimation to enable extraction of semantic guidance from VFM features. Following the Depth Anything ~\cite{yang2024depth} implementation, we set the feature dimensionality to 256 with \texttt{dptoutchannels = [128, 256, 512, 512]}. Models are trained for 100 epochs with batch size 128 (16 $\times$ 8 GPUs), learning rate 0.0016, AdamW optimizer with 10-epoch linear warmup, and weight decay 0.0001. For semantic segmentation, we use polynomial scheduling with cross-entropy loss (19 classes); for depth estimation, cosine annealing with cross-entropy loss (256 classes).

\section{Limitations and Future Work}

While the proposed \ours framework already demonstrates strong semantic consistency, efficient training, and robust generative performance across multiple scenarios, it naturally opens several promising avenues for further exploration. A key opportunity lies in broadening the diversity and granularity of visual foundation model (VFM) features used for guidance. Incorporating richer 3D perception cues, scene-level geometry, or collaborative encoders such as Radiov2.5~\cite{heinrich2025radiov2} or DUNE~\cite{sariyildiz2025dune} could provide more structured semantic priors and enhance multimodal reasoning. In parallel, integrating explicit controllability mechanisms—such as text prompts, trajectory constraints, or high-level scene graphs—may transform Re2Pix into a more general conditional video generation interface capable of supporting a wide range of user-driven editing and synthesis tasks.


\section{Broader Impact}

Our approach has the potential to benefit a wide range of societal applications by making semantic video prediction more accessible, adaptable, and robust. By forecasting future frames through semantically meaningful vision foundation model features, our method enables flexible deployment across diverse decision-making tasks—such as urban autonomy, robotics, and infrastructure monitoring—without requiring costly retraining or domain-specific adaptation. While we do not anticipate direct risks from the methodology itself, we acknowledge that the quality, reliability, and fairness of predictions ultimately depend on the pretraining data and potential biases inherited from large vision foundation models. Careful evaluation and responsible model use remain essential when applying such systems in high-stakes environments.



\begin{figure*}[t!]
    \centering
    \include{figs/fig_supp_visual_228}
    \vspace{-20pt}
    \caption{\textbf{Qualitative example on Cityscapes (scene 228).} Top: context frames with corresponding DINOv2 features (PCA visualization). Bottom: ground-truth future frames, predictions from the baseline, and predictions from \ours with their forecasted semantic features. Red boxes highlight regions where \ours provides noticeably more accurate than the baseline.}
    \label{fig:qualitative-scene_228}
\end{figure*}

\begin{figure*}[t!]
    \centering
    \include{figs/fig_supp_visual_289}
    \vspace{-20pt}
    \caption{\textbf{Qualitative example on Cityscapes (scene 289).} Top: context frames with corresponding DINOv2 features (PCA visualization). Bottom: ground-truth future frames, predictions from the baseline, and predictions from \ours with their forecasted semantic features. Red boxes highlight regions where \ours provides noticeably more accurate than the baseline.}
    \label{fig:qualitative-scene_289}
\end{figure*}


\begin{figure*}[t!]
   \centering
   \include{figs/fig_supp_visual_456}
   \vspace{-20pt}
    \caption{\textbf{Qualitative example on Cityscapes (scene 456).} Top: context frames with corresponding DINOv2 features (PCA visualization). Bottom: ground-truth future frames, predictions from the baseline, and predictions from \ours with their forecasted semantic features. Red boxes highlight regions where \ours provides noticeably more accurate than the baseline.}
   \label{fig:qualitative-scene_456}
\end{figure*}

\begin{figure*}[t!]
    \centering
    \include{figs/fig_supp_visual_485}
    \vspace{-20pt}
    \caption{\textbf{Qualitative example on Cityscapes (scene 485).} Top: context frames with corresponding DINOv2 features (PCA visualization). Bottom: ground-truth future frames, predictions from the baseline, and predictions from \ours with their forecasted semantic features. Red boxes highlight regions where \ours provides noticeably more accurate than the baseline.}
    \label{fig:qualitative-scene_485}
\end{figure*}


\end{document}